\documentclass{article}

\usepackage{microtype}
\usepackage{graphicx}
\usepackage{subfigure}
\usepackage{booktabs} 

\usepackage{hyperref}
\usepackage{multicol,multienum, multirow}
\usepackage{amsmath}

\usepackage[accepted]{icml2021}

\begin{document}


\icmltitle{Deconvolution-and-convolution Networks}



\icmlsetsymbol{equal}{*}

\begin{icmlauthorlist}
\icmlauthor{Yimin Yang}{equal,to,goo}
\icmlauthor{Wandong Zhang}{equal,to,ed}
\icmlauthor{Jonathan Wu}{ed}
\icmlauthor{Will Zhao}{egf}
\icmlauthor{Ao Chen}{to}

\end{icmlauthorlist}

\icmlaffiliation{to}{Department of Computer Science, Lakehead University, Canada}
\icmlaffiliation{goo}{Vector Institute, Canada}
\icmlaffiliation{ed}{Department of Electrical and Computer Engineering, University of Windsor, Canada}
\icmlaffiliation{egf}{Faculty of Business Administration, Lakehead University, Canada}

\icmlcorrespondingauthor{Yimin Yang}{yyang48@lakeheadu.ca}


\vskip 0.3in



\printAffiliationsAndNotice{\icmlEqualContribution} 

\begin{abstract}
2D Convolutional neural network (CNN) has arguably become the de facto standard for computer vision tasks. Recent findings, however, suggest that CNN may not be the best option for 1D pattern recognition, especially for datasets with over $1\, M$ training samples, e.g., existing CNN-based methods for 1D signals are highly reliant on human pre-processing. Common practices include utilizing discrete Fourier transform (DFT) to reconstruct 1D signal into 2D array. To add to extant knowledge, in this paper, a novel 1D data processing algorithm is proposed for 1D big data analysis through learning a deep deconvolutional-convolutional network. Rather than resorting to human-based techniques, we employed deconvolution layers to convert 1 D signals into 2D data. On top of the deconvolution model, the data was identified by a 2D CNN. Compared with the existing 1D signal processing algorithms, DCNet boasts the advantages of less human-made inference and higher generalization performance. Our experimental results from a varying number of training patterns ($50\,K$ to $11\,M$) from classification and regression demonstrate the desirability of our new approach.
\end{abstract}

\section{Introduction}
\label{introduction}




With the fast development of science, technology and data storage, big data are now rapidly expanding in all areas and application domains. Every day, more than 306.4 billion emails are sent, and 5 million Tweets are made. By utilizing large data, more sophisticated algorithms and strategies can be proposed, and thus lead to better generalization performance~\cite{wu2013data}. The reason is simple: the larger the training dataset, the more useful insights and knowledge can be retrieved and developed. Recent 2D image image datasets, including ImageNet~\cite{deng2009imagenet} and Place-365~\cite{zhou2017places}, have already surpassed the 1 million images barrier at 1.2 and 1.8 million training patterns, respectively. At the same time, several deep 2D CNNs, such as ResNet~\cite{he2016deep} and DenseNet~\cite{huang2017densely}, have outmatched the traditional classification and regression methods in numerous real-world applications.

The same trends have occurred in 1D signal processing. For example, Baldi~\cite{baldi2014searching} developed a large 1D signal classification dataset HIGGS with a sample size of more than 11 million samples. Further, another 1D big dataset HEPMASS~\cite{baldi2016parameterized} containing more than 10 million patterns was developed in 2016. However, in contrast to the 2D pattern recognition domain where 2D CNNs immediately replaced the traditional classification methods, the reign of traditional approaches is still unchallenged for 1D signals~\cite{kiranyaz20191d}.

The existing 1D signal processing algorithms can be categorized into three categories: reconstruction-based algorithms, 1D CNN-based approaches, and reshape-based methods. Generally, the reconstruction-based algorithms~\cite{bengio2013representation} learn a direct encoding from raw input by utilizing multiple autoencoders (AEs), whereas the final label is categorized with a simple classifier. The 1D CNN-based methods~\cite{kiranyaz2015real} extract the abstractive features from a sequence of 1D data. These models have similar learning procedures as those of the 2D CNNs, but the convolutional kernels move in one direction. As for reshape-based methods~\cite{appana2017speed}, they first utilize some traditional techniques, like DFT, to reshape or represent the input 1D information into a 2D signal, and then the 2D CNNs are applied for representation learning and final classification or regression. In recent years, researchers in various real-world applications have made significant contributions to broaden the field scope of reshape-based strategies, such as fault diagnosis of bearings~\cite{lu2017intelligent} and electrocardiogram (ECG) beat classification~\cite{ruiz2018arrhythmia}. Compared to the 1D CNN-based approaches that simply utilize 1D vectors for pattern recognition, the high-dimensional information processed by reshape-based strategies are much easier to be understood and mined~\cite{ding2017energy}. Thus, in most cases, the reshape-based methods achieve superior performance over 1D CNN-based algorithms~\cite{bi546explainable}.

However, the current reshape-based algorithms cannot be directly utilized in 1D large data classification and regression because of the following two reasons: First, for the task of large data analysis, the representing of the 1D signal into the 2D pattern is difficult, due to the expensive cost associated with human-made pre-processing~\cite{kiranyaz20191d}. For example, it is inefficient to ask users to create 10 million 2D patterns using the DFT technique on the HEPMASS dataset. To address this problem, we employ a novel deconvolution model to automatically generate the 2D signal according to the input 1D data, which is more effective in handling large-scale datasets than human-made pre-processing.

Second, the existing reshape-based methods are not trained in an end-to-end manner. Besides the top performance that one CNN can achieve, one crucial advantage for CNN structure in computer vision and deep learning is that it combines feature representation learning and final pattern recognition into a single body~\cite{zhang2020deep}, reducing the human-made interference. However, the reshape-based algorithms prefer to use the traditional handcrafted method, like DFT or simply reshaping the 1D signal into the 2D matrix, to pre-process the input data~\cite{janssens2016convolutional}. Nonetheless, these strategies show remarkable performance over other 1D signal classification methods on small-scale datasets; it is not guaranteed that the same human-made pre-processing could be still effective on some large-scale datasets with more than 10 million samples. Thus, it is an intuition to develop a novel deep 1D CNN framework with purely end-to-end training. In other words, the motivation of this paper is to develop a reshape-based CNN neural network to automatically represent the 1D input into 2D data.

\begin{figure*}[!t]
\begin{minipage}[b]{1\linewidth}
\centering
\includegraphics[trim={1.0cm, 0cm, 1.5cm, 0.5cm}, clip, width=\textwidth]{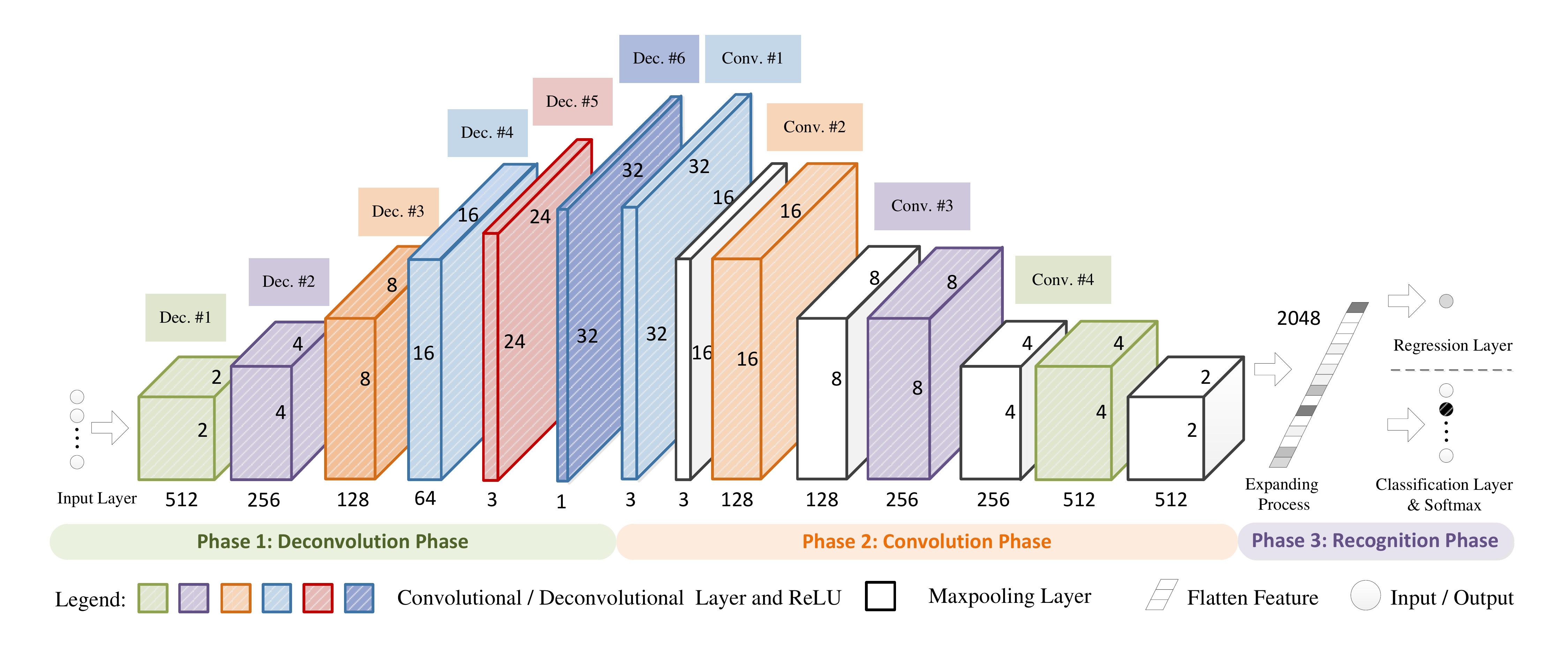}
\end{minipage}
\vspace{-0.8cm}
\caption{Flow chart of the proposed DCNet. The DCNet is composed of three learning phases, deconvolution, convolution, and recognition phases. The deconvolution phase converts the 1D signal into 2D data, and the information is identified and recognized by a 2D CNN in the convolution and recognition phases. }
\label{asa}
\end{figure*}

Driven by the mentioned motivations, we propose a completely different strategy called DCNet to perform 1D signal classification and regression using deconvolution and convolution structure. Figure~\ref{asa} shows the learning process of the proposed framework. This paper contributes the following:
\vspace{-0.1 cm}
\begin{enumerate}
    \item \textbf{\textit{A novel structure DCNet}:} We architect a deep neural network that is composed of deconvolution layers and convolution layers. Utilizing deconvolution layers to reconstruct 2D signal is meaningful, but as far as we know, no one has attempted to do so.
    \vspace{-0.1 cm}
    \item \textbf{\textit{Excellent robustness w.r.t. network structure}:} This paper demonstrates the network robustness by showing that, with the random number of deconvolutional layers (randomly selected from 3 to 6), the performance achieved by the proposed DCNet is superior to that of the existing 1D signal classification and regression models.
    \item \textbf{\textit{Remarkable performance on big data analysis}:} We achieve outstanding performance using the proposed DCNet on large datasets, such as HIGGS (w/ more than $11\, M$ patterns) and HEPMASS (w/ more than $10\, M$ samples) datasets.
\end{enumerate}


\section{Related work}

Recent years have witnessed the power of deep 2D CNN for visual data analysis, like Image classification~\cite{he2016deep}, semantic segmentation~\cite{noh2015learning}, and super-resolution imaging~\cite{umehara2018application}. The 2D CNN-based algorithms have achieved state-of-the-art and immediately outmatched the traditional recognition strategies in many real-world applications. One of the primary reasons is that CNNs are capable of acquiring the knowledge and the hidden patterns of the data automatically~\cite{zhang2020deep}, reducing the demand for human-made pre-processing.

Motivated by the strong capability of deep learning (DL) algorithms in the application of representation learning and feature extraction, researchers have utilized DL in 1D signal processing. One utilization of DL structure in this area is that of the reshape-based methods, where a proper 1D to 2D conversion is needed~\cite{kiranyaz20191d}. For example, in fault diagnosis of bearings, various techniques have been used to represent the 1D vibration signals into 2D. In~\cite{janssens2016convolutional}, the DFT is used to transform the signal from the original temporal domain to the frequency domain, and then it is represented in a 2D array which is further fed into a 2D CNN. In~\cite{zhang2017bearings}, the 1D signal is directly reshaped into a 2D vibration image. Ding \textit{et al}~\cite{ding2017energy} proposed a novel algorithm for bearing fault diagnosis through the use of wavelet packet energy (WPE) image. In this method, the phase space reconstruction is utilized to construct the 2D image. Similarly, in ECG beat classification, one widely used technique~\cite{ruiz2018arrhythmia, zihlmann2017convolutional} in this area is to represent the ECG signal into 2D images by computing power- or log-spectrogram. Compared to the other 1D signal classification algorithms, such as the reconstruct-based~\cite{wu2020fault} and 1D CNN-based classification methods~\cite{kiranyaz2015real, hannun2019sequence}, the reshape-based strategies have the advantages of excellent robustness and better generalization performance.

Though the existing reshape-based strategies show remarkable performance on small-scale 1D signal classification and regression, they cannot be directly utilized in large-scale data analysis due to two limitations. First, they are only verified on small datasets. For example, in Ding's work~\cite{ding2017energy}, the proposed reshape-based model is validated on a small dataset consisting of 100 patterns. Similarly, authors in~\cite{ruiz2018arrhythmia} only focused on processing a small dataset with 2,048 samples. Thus, the reshape-based algorithms for big data analysis have not been proposed.

Second, the existing reshape-based methods utilize two separate stages for 2D signal representation and final pattern classification, which increases the effort of human design. Researchers~\cite{zhang2020width, valmadre2017end} have already verified that the generalization performance of a model trained from several designed processes cannot be expected to be perfectly aligned with end-to-end learning. The success of recent proposed 2D CNNs indicates that the generalization performance of one model can be much better if the network is learned from the rawest possible data to the final output~\cite{levine2016end}.

To overcome these issues, this research develops a new 1D data classification algorithm called DCNet. The experimental results on large datasets showcase the effectiveness of the proposed strategy.

\section{Proposed DCNet Architecture}

\subsection{Network Formation}

The strategy to model the network is based on deconvolution and convolution operation. The former part aims to represent the 1D signal into 2D data, while the latter part prefers to identify (classify or predict) the reshaped 2D pattern.

\subsubsection{Purposes} The purpose of the proposed algorithm is to have a reshape-based algorithm for 1D signal processing by purely end-to-end learning. The DCNet is to be loaded with 1D data, and it should automatically represent each 1D input pattern into the 2D matrix. Further, the input signal is identified (classification or regression) by a 2D CNN. The whole process needs to be finished in an end-to-end manner, which means the algorithm must be sufficiently accurate that do not need pre-processing and post-processing.

\subsubsection{The structure} The structure of the proposed DCNet is depicted as Fig.~\ref{asa}. Table~\ref{t1} provides details about the layer connectivity types, kernel sizes, and output dimensions. The proposed DCNet consists of two phases: the first part is the deconvolution phase for 2D data representation; the second one is the convolution phase corresponds to the feature extractor for dimension reduction and pattern recognition. According to different pattern recognition purposes (classification and regression), the final layer is connected with a Softmax classifier or a predictor.

\begin{table}
\small
\centering
\vspace{-0.2 cm}
\caption{Layer details of DCNet}
\vspace{0.2 cm}
\setlength\tabcolsep{6.0pt}
\begin{tabular}{l|l|l|lllcccccccc}
\toprule
&\multirow{2}*{Layer ID}  &\,\,\,\,\,\,\,\,\,\,\,\,\,Layer Type &\,\,\,\,\,\,\,\,\,\,\,\,\,Output Shape\\
 &&\,\,\,\,\,\,\,\,\,\,\,\,\,\,\,\,A(\textit{k}, \textit{s})  &\,\,\,\,\,\,\,\,\,\,\,\,\,\,\,\,[\textit{b}, \textit{h}, \textit{w}, \textit{c}]\\
\midrule
\multirow{12}*{\rotatebox{90}{Deconvolution Phase}}          & Input   & Input Layer                              & \,\,\,\,\,\,\,\,[\textit{b}, 1, 1, \textit{c}]   \\
& L1      & Deconv (2, 1)                             &\,\,\,\,\,\,\,\,[\textit{b}, 2, 2, 512]   \\
& L2      & ReLU Layer                                     &\,\,\,\,\,\,\,\,[\textit{b}, 2, 2, 512]\\
& L3      & Deconv (3, 1)                             &\,\,\,\,\,\,\,\,[\textit{b}, 4, 4, 256]\\
& L4      & ReLU Layer                                 &\,\,\,\,\,\,\,\,[\textit{b}, 4, 4, 256]\\
& L5      & Deconv (5, 1)                             &\,\,\,\,\,\,\,\,[\textit{b}, 8, 8, 128]\\
& L6      & ReLU Layer                                 &\,\,\,\,\,\,\,\,[\textit{b}, 8, 8,128]\\
& L7      & Deconv (9, 1)                             &\,\,\,\,\,\,\,\,[\textit{b}, 16, 16, 64]\\
& L8      & ReLU Layer                                &\,\,\,\,\,\,\,\,[\textit{b}, 16, 16, 64]\\
& L9      & Deconv (9, 1)                             &\,\,\,\,\,\,\,\,[\textit{b}, 24, 24, 3]\\
& L10      & ReLU Layer                                &\,\,\,\,\,\,\,\,[\textit{b}, 24, 24, 3]\\
& L11      & Deconv (9, 1)                             &\,\,\,\,\,\,\,\,[\textit{b}, 32, 32, 1]\\
& L12      & ReLU Layer                                &\,\,\,\,\,\,\,\,[\textit{b}, 32, 32, 1]\\
\midrule
\multirow{8}*{\rotatebox{90}{Convolution Phase}}
& L13      & Conv (2, 1)           & \,\,\,\,\,\,\,\,[b, 32, 32, 3]\\
& L14      & ReLU $\rightarrow$ Maxpooling                        & \,\,\,\,\,\,\,\,[\textit{b}, 16, 16, 3]\\
& L15      & Conv (2, 1)           & \,\,\,\,\,\,\,\,[\textit{b}, 16, 16, 128]\\
& L16      & ReLU $\rightarrow$ Maxpooling                        & \,\,\,\,\,\,\,\,[\textit{b}, 8, 8, 128]\\
& L17      & Conv (2, 1)           & \,\,\,\,\,\,\,\,[\textit{b}, 8, 8, 256]\\
& L18      & ReLU $\rightarrow$ Maxpooling                        & \,\,\,\,\,\,\,\,[\textit{b}, 4, 4, 256]\\
& L19      & Conv (2, 1)           & \,\,\,\,\,\,\,\,[\textit{b}, 4, 4, 512]\\
& L20      & ReLU $\rightarrow$ Maxpooling                        & \,\,\,\,\,\,\,\,[\textit{b}, 2, 2, 512]\\
\midrule
& L21      & Fully-connected                         &\,\,\,\,\,\,\,\,[\textit{b}, 1, 1, \textit{l}]\\
\midrule
\multicolumn{4}{c}{A(\textit{k}, \textit{s}): A - Type of operation, \textit{k} - kernal size, \textit{s} - stride}
\\
\multicolumn{4}{c}{[\textit{b}, \textit{h}, \textit{w}, \textit{c}]: \textit{b} - batch size, \textit{h} - height, \textit{w} - weight, \textit{c} - channels}
\\
\bottomrule
\end{tabular}
\label{t1}
\end{table}

\textbf{I. Deconvolution phase}. This phase contains five deconvolutional layers. It starts from the input layer to the L9 layer, where the dimension size equal to 32$\times$32. However, as for the input dimension, the dimensionality of the input layer is determined based on the processing dataset. For example, for Connect4~\cite{burton2006performance} dataset, the input feature of each pattern is 126. Thus, the input layer dimension for DCNet on Connect4 set is $1\times1\times126$ (The input size is $1\times1$ and the input channel is 126). Similarly, for HIGGS~\cite{baldi2014searching} dataset, the dimension for each 1D pattern is 28. Therefore, the input layer channel for DCNet on this dataset is 26. In contrast to the convolution phase for which the size of feature reduces through feedforwarding, in the proposed DCNet, the dimension of features steadily increases by the use of deconvolutional layers.

\textbf{II. Convolution phase}. This phase is composed of four convolutional layers beginning from the Conv layer (L13) to the max-pooling layer (L20). In particular, for the classification, the final layer (L21) is a fully-connected layer with a softmax function. As for the regression task, the last output layer is a predictor with only one neuron. Similar to the other CNNs, the convolution phase of DCNet not only has the convolutional layers, but also contains ReLU layers and max-pooling layers.

\begin{figure}[!t]
\centering
\includegraphics[trim={1.0cm, 0.5cm, 1.0cm, 0.5cm}, clip, width=0.5\textwidth]{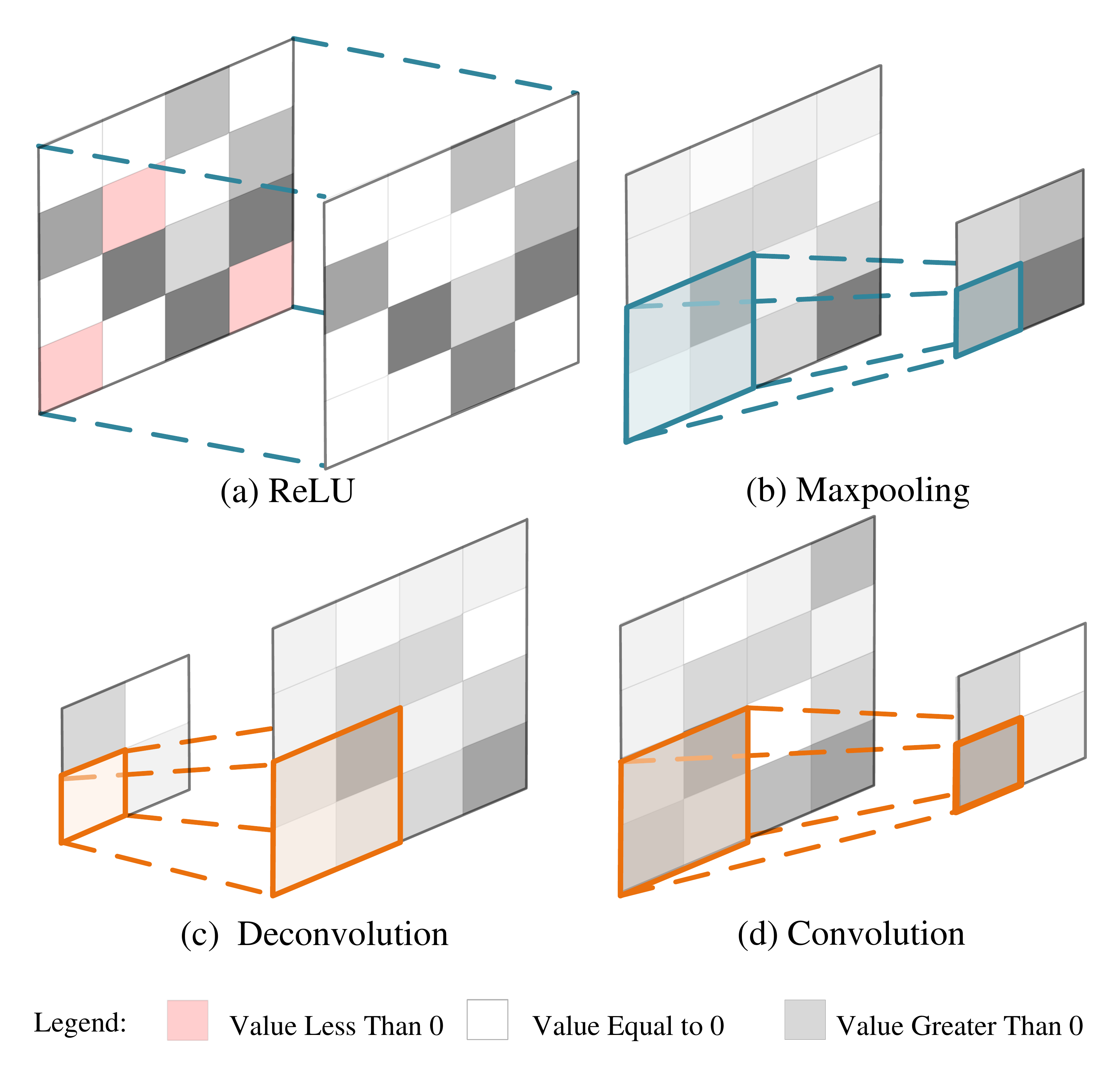}
\vspace{-0.8 cm}
\caption{Illustration of ReLU, max-pooling, deconvolution and convolution operations.}
\label{a1}
\end{figure}


\subsection{Deconvolutional Layers}

The deconvolution or transposed convolution operation is an up-sampling process that both up-samples feature maps and maintain the connectivity pattern. Essentially, the deconvolutional layers enlarge and densify the input by utilizing convolution-like operations with multiple filters. In contrast to the existing resizing techniques, the deconvolution contains trainable parameters, as illustrated in Fig.~\ref{a1}. The weights of deconvolutional layers keep updating and refining during network training. It is achieved by adding zeros between the consecutive neurons in the receptive field at the input side, and then one convolution kernel is utilized on the top with unit stride~\cite{akilan2019video}.

\subsection{2D Convolutional Layers}

The convolutional layers apply convolution operation for representation learning and feature extraction, where the weights of each filter operates like a dictionary of feature pattern. For the input patch $\mathcal X$ and the filter kernel $\mathcal K$, the convolution operation can be described as the following
\begin{equation}
C(m,n)=\sum_{p=0}^{P}\sum_{q=0}^{Q}\mathcal K(p,q)*\mathcal X(m+p,n+q)+b,
\end{equation}
where $*$ is the convolution operation, \{$m,n$\}, \{$P,Q$\}, \{$p,q$\} refer to the location of 2D patch, the size of convolution kernel, and the index of convolution kernel, respectively. The DCNet rectifies the output of convolution operation through ReLU function defined as $f(x)=max(x,0)$, where $x$ is the output of convolutional layer.

The proposed DCNet is capable of processing regression and classification tasks. For the classification, a fully-connected layer is connected at the top of convolutional layers. As for the regression tasks, a prediction layer with a single neuron is developed as L21.

\subsection{Optimizer}

The proposed DCNet is trained through the stochastic gradient descent (SGD)-based optimizer. According to different application requirements, the loss function could be different. For the classification task, training the DCNet is to minimize the following cross-entropy loss function:
\begin{equation}
\begin{split}
    L&=-\frac{1}{N\times M}\sum_{n=1}^{N}\sum_{m=1}^{M}\\
    &\times \left[p_{nm}\,{\rm log}\,\hat p_{nm}+(1-p_{nm})\,{\rm log}\,(1-\hat p_{nm})\right],\\
\end{split}
\end{equation}
where $N$ is the number of training samples in one mini-batch, $M$ refers to the output layer dimension of each sample, $\hat p_{nm}$ is the output of $m$-th location in $n$-th sample, $p_{nm}$ and $\hat p_{nm}$ stand for the target and network output respectively. As for the regression task, the loss function for DCNet is:
\begin{equation}
    L=-\frac{1}{N}\sum_{n=1}^{N}(y_n-\hat y_n)^2,
\end{equation}
where $y_n$ and $\hat y_n$ are the target output and prediction for response $n$ respectively.

\begin{table}
\centering
\vspace{-0.2 cm}
\caption{Summary of the datasets}
\vspace{0.2 cm}
\setlength\tabcolsep{4.0pt}
\begin{tabular}{lcrrrccccccccc}
\toprule
Datasets  &Classes &Training  &Testing  &Attribute\\
\midrule
\multicolumn{5}{c}{\textbf{\textit{Classification Datasets}}}\\
\midrule
Connect4     &3  &50,000    & 17,557        & 126\\
Covertype    &2  &300,000   & 280,000       & 54\\
Hep-OS       &2  &640,000   & 160,000       & 27\\
Hep-NS       &2  &640,000   & 160,000       & 27\\
Hep-AS       &2  &640,000   & 160,000       & 27\\
HIGGS-S      &2  &640,000   & 160,000       & 28\\
Hep-O        &2  &7,000,000 & 3,500,000     & 27\\
Hep-N        &2  &7,000,000 & 3,500,000     & 27\\
Hep-A        &2  &7,000,000 & 3,500,000     & 27\\
HIGGS        &2  &7,700,000 & 3,300,000     & 28\\
\midrule
\multicolumn{5}{c}{\textbf{\textit{Regression Datasets}}}\\
\midrule
ethylene-ME    &1  &3,000,000 & 1,178,504   & 16\\
ethylene-CO    &1  &3,000,000 & 1,208,261   & 16\\
\bottomrule
\end{tabular}
\label{dataset}
\end{table}

When training the proposed DCNet, the initial learning rate is $1.0^{-2}$. The total training epoch is 9, with a learning rate scheduler that rate reduces the learning rate by a factor of 0.9 in every three training epochs. During training, the momentum term is set to 0.9. The mini-batch size for the small-scale dataset is set to 64. For the large-scale dataset, the size is set to 256.

\subsection{Algorithmic Summary}

The entire process of DCNet from training to inferencing can be summarized as follows:

\textbf{I. Data normalization}. For a 1D large-scale dataset, normalize the samples from their original range into [-1, 1]. Split the dataset into the training set and testing set.

\textbf{II. Training DCNet}. The training settings of DCNet is described in subsection 3.4. The weights of the model are first randomly initialized, and then they are fine-tuned jointly on one training set with different purposes (classification or regression) in an end-to-end manner.

\textbf{III. Testing DCNet}. Utilize the well-trained DCNet on the testing set and obtain the testing accuracy or the root mean square error (RMSE).

\section{Experimental Results}

In this paper, a set of experiments ranging from 1D signal classification to regression are conducted to verify the effectiveness of the proposed DCNet. Specifically, this section first describes the implementation details, then the quantitative and qualitative analysis of the proposed model is given.

\subsection{Datasets and Experimental Environment}

\subsubsection{Datasets}

We evaluate the proposed method on twelve 1D datasets. The statistics of these datasets are shown in Table~\ref{dataset}. Based on the different training fashions and application domains, the adopted datasets can be divided into three categories: small-scale classification datasets (Connect4 and Covertype), large-scale classification datasets (HEPMASS-O/Os, HEPMASS-N/Ns, HEPMASS-A/As, HIGGS and HIGGS-s), and large-scale regression datasets (ethylene-ME/CO). The small datasets are generally small samples ($<$500 $K$), whereas the large datasets have a marginally large number of samples ($\geq$500 $K$). The details of the datasets are described as follows:

\textbf{I. Small-scale classification datasets}. Following the commonly used training settings, for Connect4~\cite{burton2006performance} and Covertype~\cite{gama2003accurate} datasets, we randomly take $50\,K$ and $300\,K$ samples for training, and the rest patterns for testing.

\textbf{II. Large-scale classification datasets}. As far as we know, the HEPMASS~\cite{baldi2016parameterized} and HIGGS~\cite{baldi2014searching} datasets could be the largest sets in the 1D signal classification area. The HEPMASS contains three datasets (collected from different environments) with 7\,$M$ training samples and 3.5\,$M$ testing patterns, which are HEPMASS-O, HEPMASS-N and HEPMASS-A, respectively (We use Hep-O, Hep-N and Hep-A to denote for simplification). As for the HIGGS dataset, it has 7.7\,$M$ training samples and 3.3\,$M$ testing data. To fully evaluate the performance of the algorithms, we randomly selected 700 $K$ samples from each training set to generate Hep-OS, Hep-NS, Hep-AS, and HIGGS-S datasets (in each set 80\% samples for training while the rest for testing).

\textbf{III. Large-scale regression datasets}. In this paper, two large regression datasets~\cite{fonollosa2015reservoir} are utilized for validation, which are the ethylene-ME containing 4,178,504 samples and the ethylene-CO consisting of 4,208,261 patterns. The datasets were collected in a gas delivery platform from 16 chemical sensors for gas mixtures prediction. Specifically, the ethylene-ME and ethylene-CO datasets are collected for predicting two kinds of gas mixtures, i.e., Ethylene and Methane in air, and Ethylene and CO in air. For these two datasets, we randomly chose 3\,$M$ patterns for training the network, and the rest for testing it.

\subsubsection{Rival Methods}

We compare the proposed method with some state-of-the-art learning baseline, including:

\begin{table}
\centering
\vspace{-0.2 cm}
\caption{Top-1 testing accuracy comparison ($\%$) with different structures (Dec. - deconvolutional layers)}
\vspace{0.2 cm}
\setlength\tabcolsep{7.0pt}
\begin{tabular}{lccccccccccccc}
\toprule
\multirow{2}*{Datasets}  &DCNet w/         &DCNet w/               &DCNet w/ \\
&Three Dec.     &Six Dec.   &Random Dec.\\
\midrule
Hep-OS     &91.6    &91.8   &91.5\\
Hep-NS     &85.5    &85.9   &85.7\\
Hep-O      &92.6    &92.6   &92.6\\
Hep-N      &87.0    &87.3   &87.0\\
\bottomrule
\end{tabular}
\label{t3}
\end{table}

\textbf{WD-AE}~\cite{charte2018practical}: An autoencoder (AE) can be used for representation learning by back-propagation (BP). After feature learning, the Softmax classifier is utilized to get the final labels.

\textbf{Wi-SNN}~\cite{zhang2020width}: A deep width-growth model trained with least-square based Moore-Penrose (MP) inverse technique. This model could directly obtain the final labels.

\textbf{Random Forest}~\cite{breiman2001random}: It is an ensemble learning algorithm by constructing a multitude of decision trees during training.

\textbf{LSTM}~\cite{hochreiter1997long}: Long short-term memory (LSTM) is a recurrent neural network for time-seties data classification.

\textbf{Deep-LSTM}~\cite{sagheer2019time}: Deep LSTM is composed of several LSTM layers. It performs well on processing time-series data. In this paper, two LSTM layers are stacked to develop a deep model.

\textbf{1D-CNN}~\cite{acharya2017application}: A 1D pattern recognition network identified by1D CNN. It learns the labels of input data with four convolutional layers and two fully-connected layers.

\textbf{B-CNN}~\cite{giudice20201d}: It is a convolutional network for processing 1D EEG signals. This model achieves the state-of-the-art in EEG signal processing.

\begin{table}
\centering
\vspace{-0.2 cm}
\caption{Top-1 testing accuracy comparison ($\%$) with different size of mini-batch ($b$ - size of mini-batch)}
\vspace{0.2 cm}
\setlength\tabcolsep{8.0pt}
\begin{tabular}{lcrrrccccccccc}
\toprule
Datasets  &\textit{b} = 32 &\textit{b} = 64  &\textit{b} = 128  &\textit{b} = 256\\
\midrule
Connect4     &85.5  &85.4   &85.2   &84.2\\
Covertype    &95.1  &95.0   &94.4   &94.0\\
Hep-O        &87.3  &87.3   &87.2   &87.3\\
HIGGS        &76.0  &76.1   &76.0   &75.7\\
\bottomrule
\end{tabular}
\label{t4}
\end{table}

\subsubsection{Experimrntal Environment}

The validation experiments are conducted in MATLAB R2020a environment in a workstation with Intel Core $E5-2650$ CPU and 256 GB memory. The Top-1 testing accuracy is used for evaluating classification tasks, and the RMSE is utilized in regression. In this paper, for each algorithm, at least five trials are conducted for each dataset.

\begin{table*}[!t]
\vspace{0.1 cm}
\setlength\tabcolsep{8pt}
 \centering
\caption{Comparison of the proposed DCNet and other state-of-the-art 1D signal classification algorithms w.r.t. Top-1 testing accuracy ($\%$). Values in \textcolor{red}{RED} are the best figures for one dataset, the ones in \textcolor{blue}{BLUE} are the second best results (RNO - results are not available due to out of main memory)}
\vspace{0.2 cm}
\begin{tabular}{lccccccccccccccc}
\toprule
Datasets   & WD-AE   &Wi-SNN &Random Forest &LSTM   &Deep-LSTM  &1D-CNN     &B-CNN  &DCNet\\
\midrule
Connect4    &77.9   &75.6   &82.2   &72.5   &73.8   &\textcolor{blue}{82.3}   &81.9   &\textcolor{red}{85.4}\\
Covertype   &72.1   &74.9   &\textcolor{blue}{92.1}   &73.7   &74.9   &86.5   &81.3   &\textcolor{red}{95.2}\\
Hep-OS      &86.7   &90.8   &90.3   &89.6   &90.5   &\textcolor{blue}{90.4}   &89.9   &\textcolor{red}{91.8}\\
Hep-NS      &83.4   &81.1   &81.8   &80.2   &80.3   &\textcolor{blue}{82.5}   &82.3   &\textcolor{red}{85.9}\\
Hep-AS      &81.4   &82.7   &\textcolor{blue}{84.0}   &83.2   &81.4   &83.6   &83.8   &\textcolor{red}{86.8}\\
HIGGS-S     &63.5   &64.6   &65.4   &54.8   &57.3   &\textcolor{red}{71.7}   &64.2   &\textcolor{blue}{70.4}\\
Hep-O       &91.7   &RNO    &90.8   &91.1   &90.4   &\textcolor{blue}{91.6}   &90.6   &\textcolor{red}{92.6}\\
Hep-N       &84.9   &RNO    &\textcolor{blue}{85.1}   &84.2   &84.3   &84.5   &83.1   &\textcolor{red}{87.3}\\
Hep-A       &84.0   &RNO    &83.7   &85.2   &82.1   &84.8   &\textcolor{blue}{85.2}   &\textcolor{red}{87.2}\\
HIGGS       &69.5   &RNO    &69.6   &64.9   &67.7   &\textcolor{blue}{72.9}   &69.1   &\textcolor{red}{75.7}\\
\midrule
Average     &79.6   &RNO    &82.5   &77.9   &78.3   &\textcolor{blue}{83.1}   &81.2   &\textcolor{red}{85.7}\\
\bottomrule
\end{tabular}
\label{t5}
\end{table*}

\begin{figure*}[t]
\begin{center}
  \includegraphics[trim={1.5cm, 1cm, 1.5cm, 1cm}, clip, width=0.85\linewidth]{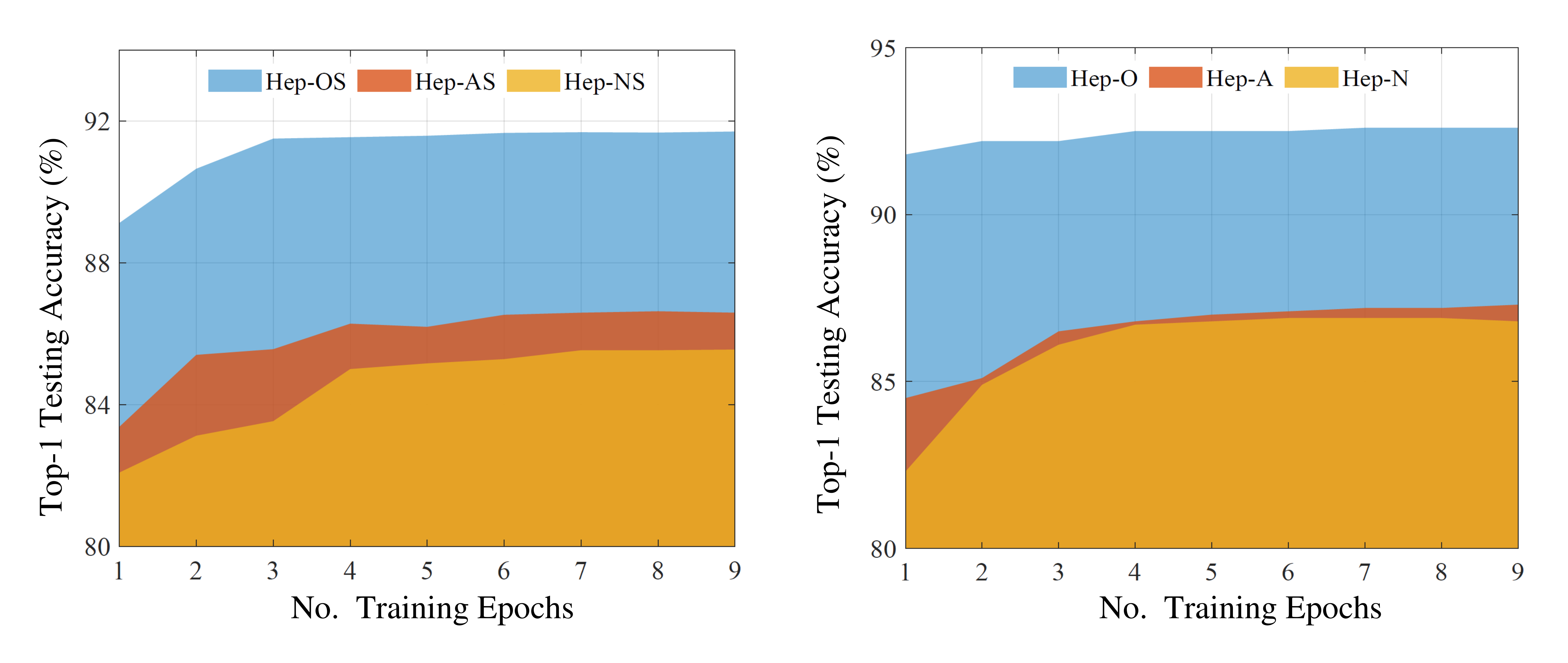}
\end{center}
\vspace{-0.5 cm}
  \caption{Top-1 testing accuracy of DCNet on Hep-OS, Hep-AS, Hep-NS, Hep-O, Hep-A and Hep-N datasets.}
\label{f3}
\end{figure*}

\subsection{Quantitative Analysis}

\subsubsection{Model Settings}

The proposed DCNet contains several hyper-parameters, such as the mini-batch size and the layer connectivity patterns. In this subsection, we verify the recommendations of these hyper-parameters.

To evaluate the impact of network structure, experiments are implemented with three types of layer connectivity patterns: the DCNet that only contains three deconvolution layers (the structure is listed in the supplementary file due to the space constraint); the DCNet containing six deconvolution layers (the network is structured as Table~\ref{t1}); the DCNet with a random number of deconvolution layers. Here the number of deconvolution layers is randomly selected from [3, 6]. In this evaluation, although the number of deconvolution layers in each condition differs, the dimensionality of the represented 2D data and the convolution part are the same for consistency. Table~\ref{t3} shows the performance of various network structures. When handling the 1D signals, the DCNet with different structures provides almost the same performance. Thus, we do not need to use trial-and-error to search for the optimal structure. For consistency and fair comparison, in this paper, the structure described in Table~\ref{t1} is utilized for comparison.

Further, to evaluate the impact of mini-batch size, a set of experiments are conducted as shown in Table~\ref{t4}. One can steadily see from Table~\ref{t4} that as the mini-batch size increases, the generalization performance of the DCNet almost remain the same. However, the large mini-batch size can lead to speeding up the network training. Thus, in this paper, for the small-scale datasets, such as Connect4, the mini-batch size is 64. As for the large-scale datasets, like Hep-O and HIGGS, the mini-batch size is set to 256.

\subsubsection{Analysis on Classification Datasets}

The overall comparison of the DCNet and other state-of-the-art 1D data processing strategies on the classification datasets are tabulated in Table~\ref{t5}. In this table, along with the testing accuracy on each dataset, the mean average performance for each algorithm is given as well. The results in Table~\ref{t5} shows that in most cases, the proposed DCNet provides superior performance than the existing models from 3\% to 10\% of improvement in Top-1 testing accuracy. For example, our results on Connect4, Hep-NS, and Hep-N datasets are 85.4\%, 85.9\%, and 87.3\% respectively; 3.1\%, 3.4\%, and 2.2\% better than the state-of-the-art results. Furthermore, the average result of the proposed method among all the datasets is 85.7\%, providing a 2.6\% and 3.2\% of accuracy improvement over the best (1D-CNN~\cite{acharya2017application}) and the second-best (random forest~\cite{breiman2001random}) algorithms in the literature. The DCNet, therefore, outmatches the other 1D classification algorithms.

Figure~\ref{f3} shows the generalization performance of DCNet by line charts as the number of learning epochs increase. Specifically, the chart on the left side is the testing accuracy on Hep-OS, Hep-AS and Hep-NS datasets, while the right chart is the performance of the proposed method on Hep-O, Hep-A and Hep-N sets. We can easily conclude that the DCNet only needs 3 to 4 learning epochs to get the optimal results. Thus, the profits of the proposed algorithm have been validated.

\begin{figure}[t]
\begin{center}
  \includegraphics[trim={1cm, 1cm, 1cm, 1cm}, clip, width=0.5\linewidth]{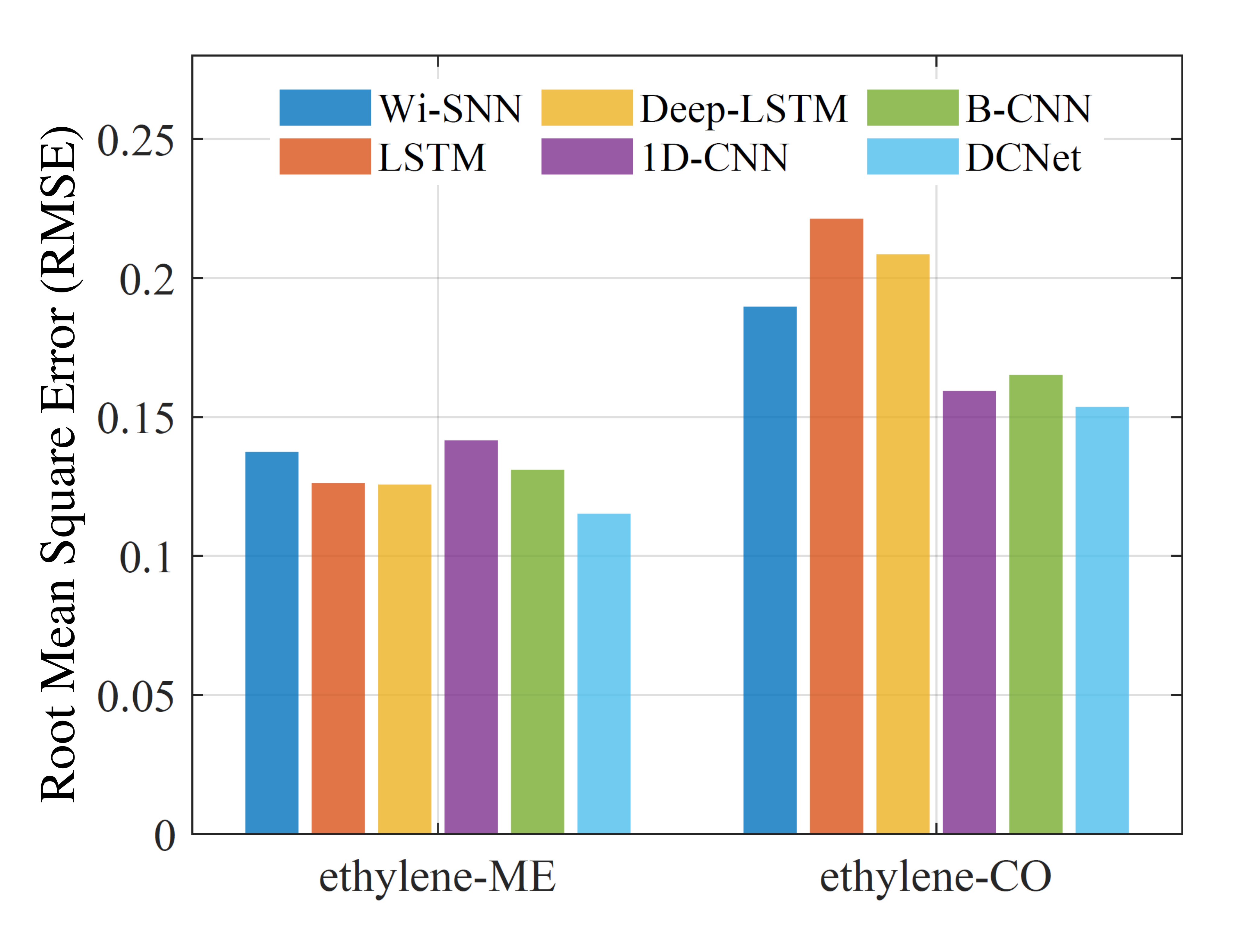}
\end{center}
\vspace{-0.7 cm}
  \caption{Performance comparison of DCNet with other algorithms on ethylene-ME and ethylene-CO datasets.}
  \vspace{-0.2 cm}
\label{f4}
\end{figure}

\subsubsection{Analysis on Regression Datasets}

The experimental results on regression tasks are shown in Fig.~\ref{f4}. It is observed that the proposed model achieves competent performance over the other 1D regression algorithms. The reason is twofold. i) The reshape-based methods convert the 1D signal into 2D data, and the information is easier to be understood in a high dimension. ii) The existing reshape-based frameworks have stringent demand for computational resources due to human-made pre-processing, which cannot be used in large data analysis. The DCNet, on the other hand, is implemented more efficiently, i.e., the 2D signal representation and pattern recognition are combined in an end-to-end framework.

\begin{figure}[hpt!]
\begin{center}
  \includegraphics[trim={1cm, 0.5cm, 1cm, 0.5cm}, clip, width=0.7\linewidth]{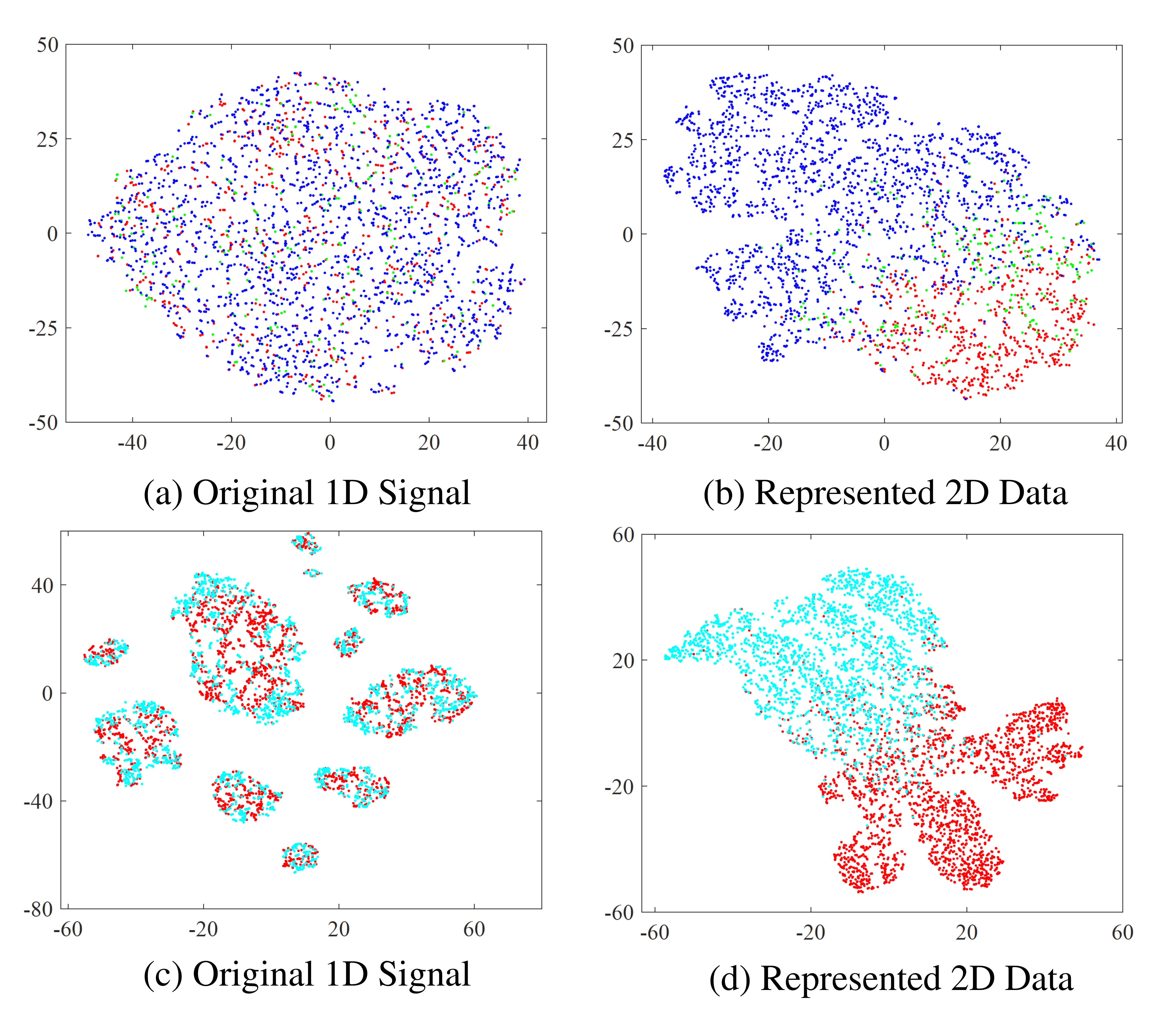}
\end{center}
\vspace{-0.6 cm}
  \caption{T-sne results of the DCNet. (a) (b) are the results of Connect4 set, and (c) (d) are the results on Hep-OS set.}
\label{f9}
\vspace{-0.2 cm}
\end{figure}

\begin{figure}[t]
\begin{center}
  \includegraphics[trim={1cm, 0.5cm, 1cm, 0.5cm}, clip, width=0.7\linewidth]{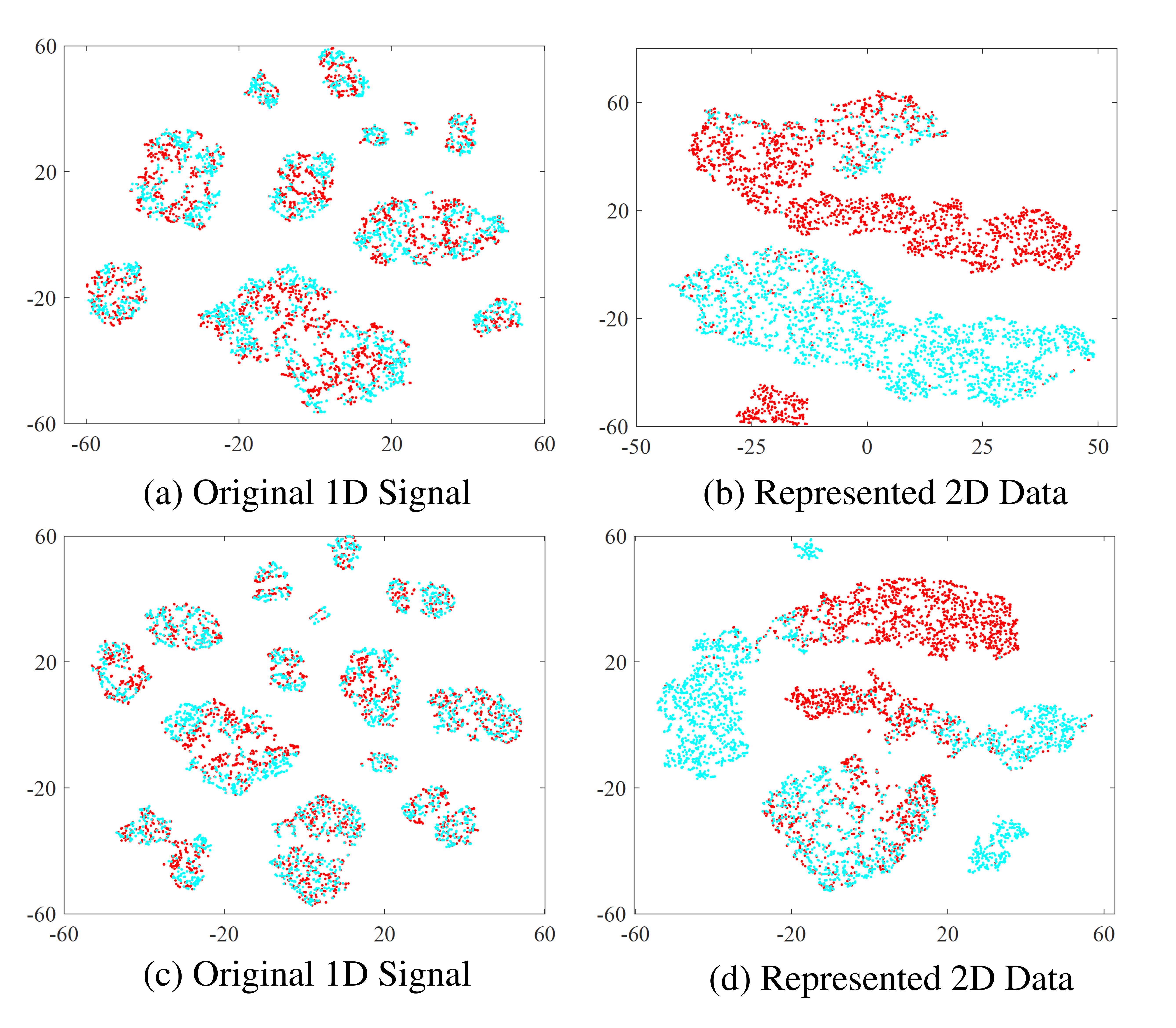}
\end{center}
\vspace{-0.6 cm}
  \caption{T-sne results of the DCNet. (a) (b) are the results of Hep-O set, and (c) (d) are the results on Hep-A set.}
  \vspace{-0.2 cm}
\label{f10}
\end{figure}

\subsection{Qualitative Analysis}

The samples visualized results with the proposed DCNet on Connect4, Hep-OS, Hep-O and Hep-A datasets are shown in Fig.~\ref{f9} and Fig.~\ref{f10}. The t-distribution stochastic neighbour embedding (t-sne) algorithm is applied to visualize the features extracted from the original input space (1D signal) and the represented 2D data (extracted from L11). Through observation, we can conclude that after the deconvolutional layers, the intra-category distances are reduced, while the inter-category distances are increased.

Taking the presented outcomes of Table~\ref{t5}, Fig.~\ref{f4} to Fig.~\ref{f10}, we can conclude that the DCNet achieves better generalization performance over other 1D processing algorithms on large-scale datasets. The reason could be that the traditional 1D signal processing algorithms apply human-made pre-processing, such as DFT, onto the 1D input to generate the 2D patches or extract the features. In fact, in some complex tasks like processing large data, there are often many more clues that can be directly encoded, the handcrafted techniques, however, by definition, cannot know exactly what is useful for a certain task. In other words, the human-made procedures cannot directly build a connection between input attributes and output targets. The proposed DCNet, on the other hand, can automatically build the proper latent space to represent the 1D raw data into the 2D feature. In other words, the DCNet is capable of learning some clues that the human-made algorithms cannot learn.

\section{Conclusion}

The paper proposes a novel architecture DCNet to effectively handle 1D signal classification and regression. The DCNet is an end-to-end training method where the data reconstruction and pattern recognition are combined in a single model. It comprises two consecutive phases: the deconvolution phase for data conversion and the convolution phase for data identification. Experimental results on large-scale datasets with more than 10 million samples validate the effectiveness of the proposed method. Compared to the existing 1D pattern processing algorithms, the DCNet is found to have the advantages of higher generalization performance and better network robustness.

\bibliography{example_paper}

\begin{thebibliography}{36}
\providecommand{\natexlab}[1]{#1}
\providecommand{\url}[1]{\texttt{#1}}
\expandafter\ifx\csname urlstyle\endcsname\relax
  \providecommand{\doi}[1]{doi: #1}\else
  \providecommand{\doi}{doi: \begingroup \urlstyle{rm}\Url}\fi

\bibitem[Acharya et~al.(2017)Acharya, Fujita, Oh, Hagiwara, Tan, and
  Adam]{acharya2017application}
Acharya, U.~R., Fujita, H., Oh, S.~L., Hagiwara, Y., Tan, J.~H., and Adam, M.
\newblock Application of deep convolutional neural network for automated
  detection of myocardial infarction using ecg signals.
\newblock \emph{Information Sciences}, 415:\penalty0 190--198, 2017.

\bibitem[Akilan et~al.(2019)Akilan, Wu, and Zhang]{akilan2019video}
Akilan, T., Wu, Q.~J., and Zhang, W.
\newblock Video foreground extraction using multi-view receptive field and
  encoder--decoder dcnn for traffic and surveillance applications.
\newblock \emph{IEEE Transactions on Vehicular Technology}, 68\penalty0
  (10):\penalty0 9478--9493, 2019.

\bibitem[Appana et~al.(2017)Appana, Ahmad, and Kim]{appana2017speed}
Appana, D.~K., Ahmad, W., and Kim, J.-M.
\newblock Speed invariant bearing fault characterization using convolutional
  neural networks.
\newblock In \emph{International Workshop on Multi-disciplinary Trends in
  Artificial Intelligence}, pp.\  189--198. Springer, 2017.

\bibitem[Baldi et~al.(2014)Baldi, Sadowski, and Whiteson]{baldi2014searching}
Baldi, P., Sadowski, P., and Whiteson, D.
\newblock Searching for exotic particles in high-energy physics with deep
  learning.
\newblock \emph{Nature communications}, 5\penalty0 (1):\penalty0 1--9, 2014.

\bibitem[Baldi et~al.(2016)Baldi, Cranmer, Faucett, Sadowski, and
  Whiteson]{baldi2016parameterized}
Baldi, P., Cranmer, K., Faucett, T., Sadowski, P., and Whiteson, D.
\newblock Parameterized neural networks for high-energy physics.
\newblock \emph{The European Physical Journal C}, 76\penalty0 (5):\penalty0
  1--7, 2016.

\bibitem[Bengio et~al.(2013)Bengio, Courville, and
  Vincent]{bengio2013representation}
Bengio, Y., Courville, A., and Vincent, P.
\newblock Representation learning: A review and new perspectives.
\newblock \emph{IEEE Transactions on Pattern Analysis and Machine
  Intelligence}, 35\penalty0 (8):\penalty0 1798--1828, 2013.

\bibitem[Bi et~al.()Bi, Zhang, He, Zhao, Sun, and Ma]{bi546explainable}
Bi, X., Zhang, C., He, Y., Zhao, X., Sun, Y., and Ma, Y.
\newblock Explainable time--frequency convolutional neural network for
  microseismic waveform classification.
\newblock \emph{Information Sciences}, 546:\penalty0 883--896.

\bibitem[Breiman(2001)]{breiman2001random}
Breiman, L.
\newblock Random forests.
\newblock \emph{Machine Learning}, 45\penalty0 (1):\penalty0 5--32, 2001.

\bibitem[Burton \& Kelly(2006)Burton and Kelly]{burton2006performance}
Burton, A.~N. and Kelly, P.~H.
\newblock Performance prediction of paging workloads using lightweight tracing.
\newblock \emph{Future Generation Computer Systems}, 22\penalty0 (7):\penalty0
  784--793, 2006.

\bibitem[Charte et~al.(2018)Charte, Charte, Garc{\'\i}a, del Jesus, and
  Herrera]{charte2018practical}
Charte, D., Charte, F., Garc{\'\i}a, S., del Jesus, M.~J., and Herrera, F.
\newblock A practical tutorial on autoencoders for nonlinear feature fusion:
  Taxonomy, models, software and guidelines.
\newblock \emph{Information Fusion}, 44:\penalty0 78--96, 2018.

\bibitem[Deng et~al.(2009)Deng, Dong, Socher, Li, Li, and
  Fei-Fei]{deng2009imagenet}
Deng, J., Dong, W., Socher, R., Li, L.-J., Li, K., and Fei-Fei, L.
\newblock Imagenet: A large-scale hierarchical image database.
\newblock In \emph{2009 IEEE Conference on Computer Vision and Pattern
  Recognition}, pp.\  248--255. Ieee, 2009.

\bibitem[Ding \& He(2017)Ding and He]{ding2017energy}
Ding, X. and He, Q.
\newblock Energy-fluctuated multiscale feature learning with deep convnet for
  intelligent spindle bearing fault diagnosis.
\newblock \emph{IEEE Transactions on Instrumentation and Measurement},
  66\penalty0 (8):\penalty0 1926--1935, 2017.

\bibitem[Fonollosa et~al.(2015)Fonollosa, Sheik, Huerta, and
  Marco]{fonollosa2015reservoir}
Fonollosa, J., Sheik, S., Huerta, R., and Marco, S.
\newblock Reservoir computing compensates slow response of chemosensor arrays
  exposed to fast varying gas concentrations in continuous monitoring.
\newblock \emph{Sensors and Actuators B: Chemical}, 215:\penalty0 618--629,
  2015.

\bibitem[Gama et~al.(2003)Gama, Rocha, and Medas]{gama2003accurate}
Gama, J., Rocha, R., and Medas, P.
\newblock Accurate decision trees for mining high-speed data streams.
\newblock In \emph{Proceedings of the Ninth ACM SIGKDD International Conference
  on Knowledge Discovery and Data Mining}, pp.\  523--528, 2003.

\bibitem[Giudice et~al.(2020)Giudice, Varone, Ieracitano, Mammone, Bruna,
  Tomaselli, and Morabito]{giudice20201d}
Giudice, M.~L., Varone, G., Ieracitano, C., Mammone, N., Bruna, A.~R.,
  Tomaselli, V., and Morabito, F.~C.
\newblock 1d convolutional neural network approach to classify voluntary eye
  blinks in eeg signals for bci applications.
\newblock In \emph{2020 International Joint Conference on Neural Networks},
  pp.\  1--7. IEEE, 2020.

\bibitem[Hannun et~al.(2019)Hannun, Lee, Xu, and Collobert]{hannun2019sequence}
Hannun, A., Lee, A., Xu, Q., and Collobert, R.
\newblock Sequence-to-sequence speech recognition with time-depth separable
  convolutions.
\newblock \emph{arXiv preprint arXiv:1904.02619}, 2019.

\bibitem[He et~al.(2016)He, Zhang, Ren, and Sun]{he2016deep}
He, K., Zhang, X., Ren, S., and Sun, J.
\newblock Deep residual learning for image recognition.
\newblock In \emph{Proceedings of the IEEE Conference on Computer Vision and
  Pattern Recognition}, pp.\  770--778, 2016.

\bibitem[Hochreiter \& Schmidhuber(1997)Hochreiter and
  Schmidhuber]{hochreiter1997long}
Hochreiter, S. and Schmidhuber, J.
\newblock Long short-term memory.
\newblock \emph{Neural computation}, 9\penalty0 (8):\penalty0 1735--1780, 1997.

\bibitem[Huang et~al.(2017)Huang, Liu, Van Der~Maaten, and
  Weinberger]{huang2017densely}
Huang, G., Liu, Z., Van Der~Maaten, L., and Weinberger, K.~Q.
\newblock Densely connected convolutional networks.
\newblock In \emph{Proceedings of the IEEE Conference on Computer Vision and
  Pattern Recognition}, pp.\  4700--4708, 2017.

\bibitem[Janssens et~al.(2016)Janssens, Slavkovikj, Vervisch, Stockman,
  Loccufier, Verstockt, Van~de Walle, and
  Van~Hoecke]{janssens2016convolutional}
Janssens, O., Slavkovikj, V., Vervisch, B., Stockman, K., Loccufier, M.,
  Verstockt, S., Van~de Walle, R., and Van~Hoecke, S.
\newblock Convolutional neural network based fault detection for rotating
  machinery.
\newblock \emph{Journal of Sound and Vibration}, 377:\penalty0 331--345, 2016.

\bibitem[Kiranyaz et~al.(2015)Kiranyaz, Ince, and Gabbouj]{kiranyaz2015real}
Kiranyaz, S., Ince, T., and Gabbouj, M.
\newblock Real-time patient-specific ecg classification by 1-d convolutional
  neural networks.
\newblock \emph{IEEE Transactions on Biomedical Engineering}, 63\penalty0
  (3):\penalty0 664--675, 2015.

\bibitem[Kiranyaz et~al.(2019)Kiranyaz, Avci, Abdeljaber, Ince, Gabbouj, and
  Inman]{kiranyaz20191d}
Kiranyaz, S., Avci, O., Abdeljaber, O., Ince, T., Gabbouj, M., and Inman, D.~J.
\newblock 1d convolutional neural networks and applications: A survey.
\newblock \emph{arXiv preprint arXiv:1905.03554}, 2019.

\bibitem[Levine et~al.(2016)Levine, Finn, Darrell, and Abbeel]{levine2016end}
Levine, S., Finn, C., Darrell, T., and Abbeel, P.
\newblock End-to-end training of deep visuomotor policies.
\newblock \emph{The Journal of Machine Learning Research}, 17\penalty0
  (1):\penalty0 1334--1373, 2016.

\bibitem[Lu et~al.(2017)Lu, Wang, and Zhou]{lu2017intelligent}
Lu, C., Wang, Z., and Zhou, B.
\newblock Intelligent fault diagnosis of rolling bearing using hierarchical
  convolutional network based health state classification.
\newblock \emph{Advanced Engineering Informatics}, 32:\penalty0 139--151, 2017.

\bibitem[Noh et~al.(2015)Noh, Hong, and Han]{noh2015learning}
Noh, H., Hong, S., and Han, B.
\newblock Learning deconvolution network for semantic segmentation.
\newblock In \emph{Proceedings of the IEEE International Conference on Computer
  Vision}, pp.\  1520--1528, 2015.

\bibitem[Ruiz et~al.(2018)Ruiz, P{\'e}rez, and
  Bl{\'a}zquez]{ruiz2018arrhythmia}
Ruiz, J.~T., P{\'e}rez, J. D.~B., and Bl{\'a}zquez, J. R.~B.
\newblock Arrhythmia detection using convolutional neural models.
\newblock In \emph{International Symposium on Distributed Computing and
  Artificial Intelligence}, pp.\  120--127. Springer, 2018.

\bibitem[Sagheer \& Kotb(2019)Sagheer and Kotb]{sagheer2019time}
Sagheer, A. and Kotb, M.
\newblock Time series forecasting of petroleum production using deep lstm
  recurrent networks.
\newblock \emph{Neurocomputing}, 323:\penalty0 203--213, 2019.

\bibitem[Umehara et~al.(2018)Umehara, Ota, and Ishida]{umehara2018application}
Umehara, K., Ota, J., and Ishida, T.
\newblock Application of super-resolution convolutional neural network for
  enhancing image resolution in chest ct.
\newblock \emph{Journal of Digital Imaging}, 31\penalty0 (4):\penalty0
  441--450, 2018.

\bibitem[Valmadre et~al.(2017)Valmadre, Bertinetto, Henriques, Vedaldi, and
  Torr]{valmadre2017end}
Valmadre, J., Bertinetto, L., Henriques, J., Vedaldi, A., and Torr, P.~H.
\newblock End-to-end representation learning for correlation filter based
  tracking.
\newblock In \emph{Proceedings of the IEEE Conference on Computer Vision and
  Pattern Recognition}, pp.\  2805--2813, 2017.

\bibitem[Wu et~al.(2020)Wu, Zhao, Sun, Yan, and Chen]{wu2020fault}
Wu, J., Zhao, Z., Sun, C., Yan, R., and Chen, X.
\newblock Fault-attention generative probabilistic adversarial autoencoder for
  machine anomaly detection.
\newblock \emph{IEEE Transactions on Industrial Informatics}, 16\penalty0
  (12):\penalty0 7479--7488, 2020.

\bibitem[Wu et~al.(2013)Wu, Zhu, Wu, and Ding]{wu2013data}
Wu, X., Zhu, X., Wu, G.-Q., and Ding, W.
\newblock Data mining with big data.
\newblock \emph{IEEE transactions on knowledge and data engineering},
  26\penalty0 (1):\penalty0 97--107, 2013.

\bibitem[Zhang et~al.(2017)Zhang, Peng, and Li]{zhang2017bearings}
Zhang, W., Peng, G., and Li, C.
\newblock Bearings fault diagnosis based on convolutional neural networks with
  2-d representation of vibration signals as input.
\newblock In \emph{MATEC web of conferences}, volume~95, pp.\  13001. EDP
  Sciences, 2017.

\bibitem[Zhang et~al.(2020{\natexlab{a}})Zhang, Wu, Yang, Akilan, and
  Zhang]{zhang2020width}
Zhang, W., Wu, Q.~J., Yang, Y., Akilan, T., and Zhang, H.
\newblock A width-growth model with subnetwork nodes and refinement structure
  for representation learning and image classification.
\newblock \emph{IEEE Transactions on Industrial Informatics},
  2020{\natexlab{a}}.

\bibitem[Zhang et~al.(2020{\natexlab{b}})Zhang, Yang, and Wu]{zhang2020deep}
Zhang, W., Yang, Y., and Wu, J.
\newblock Deep networks with fast retraining.
\newblock \emph{arXiv preprint arXiv:2008.07387}, 2020{\natexlab{b}}.

\bibitem[Zhou et~al.(2017)Zhou, Lapedriza, Khosla, Oliva, and
  Torralba]{zhou2017places}
Zhou, B., Lapedriza, A., Khosla, A., Oliva, A., and Torralba, A.
\newblock Places: A 10 million image database for scene recognition.
\newblock \emph{IEEE Transactions on Pattern Analysis and Machine
  Intelligence}, 40\penalty0 (6):\penalty0 1452--1464, 2017.

\bibitem[Zihlmann et~al.(2017)Zihlmann, Perekrestenko, and
  Tschannen]{zihlmann2017convolutional}
Zihlmann, M., Perekrestenko, D., and Tschannen, M.
\newblock Convolutional recurrent neural networks for electrocardiogram
  classification.
\newblock In \emph{2017 Computing in Cardiology (CinC)}, pp.\  1--4. IEEE,
  2017.

\end{thebibliography}
\bibliographystyle{icml2021}

\end{document}